\documentclass[letterpaper]{article} 
\usepackage{aaai23}  
\usepackage{times}  
\usepackage{helvet}  
\usepackage{courier}  
\usepackage[hyphens]{url}  
\usepackage{graphicx} 
\usepackage{natbib}  
\usepackage{caption} 
\usepackage{algorithm}
\usepackage{algorithmic}

\usepackage{xcolor}
\usepackage{amsmath}
\usepackage{pifont}
\usepackage{graphicx}

\usepackage{overpic}

\usepackage{amssymb}
\usepackage{booktabs}

\usepackage{multirow}
\newcommand{\tabincell}[2]{\begin{tabular}{@{}#1@{}}#2\end{tabular}}
\usepackage{makecell}

\usepackage{balance}

\usepackage{colortbl}
\definecolor{mygreen}{RGB}{239,255,246}

\usepackage{bm}

\usepackage{newfloat}
\usepackage{listings}
\DeclareCaptionStyle{ruled}{labelfont=normalfont,labelsep=colon,strut=off} 
\floatstyle{ruled}
\newfloat{listing}{tb}{lst}{}
\floatname{listing}{Listing}
\title{Rethinking Data-Free Quantization as a Zero-Sum Game}
\author{
    Biao Qian,
    Yang Wang\thanks{Yang Wang is the corresponding author.},
    Richang Hong,
    Meng Wang
}
\affiliations{
    Key Laboratory of Knowledge Engineering with Big Data, Ministry of Education,\\

    School of Computer Science and Information Engineering,\\
    Hefei University of Technology, China\\
    yangwang@hfut.edu.cn, \{hfutqian,hongrc.hfut,eric.mengwang\}@gmail.com
}

\usepackage{bibentry}

\begin{document}

\maketitle

\begin{abstract}
Data-free quantization (DFQ) recovers the performance of quantized network (Q) without accessing the real data, but generates the fake sample via a generator (G) by learning from full-precision network (P) instead. However, such sample generation process is totally independent of Q, specialized as failing to consider the adaptability of the generated samples, i.e., beneficial or adversarial, over the learning process of Q, resulting into non-ignorable performance loss. Building on this, several crucial questions --- how to measure and exploit the sample adaptability to Q under varied bit-width scenarios? how to generate the samples with desirable adaptability to benefit the quantized network? --- impel us to revisit DFQ. In this paper, we answer the above questions from a game-theory perspective to specialize DFQ as a zero-sum game between two players --- a generator and a quantized network, and further propose an \textbf{Ada}ptability-aware \textbf{S}ample \textbf{G}eneration (AdaSG) method. Technically, AdaSG reformulates DFQ as a dynamic maximization-vs-minimization game process anchored on the sample adaptability. The maximization process aims to generate the sample with desirable adaptability, such sample adaptability is further reduced by the minimization process after calibrating Q for performance recovery. The Balance Gap is defined to guide the stationarity of the game process to maximally benefit Q. The theoretical analysis and empirical studies verify the superiority of AdaSG over the state-of-the-arts. \emph{Our code is available at https://github.com/hfutqian/AdaSG}.
\end{abstract}

\section{Introduction}
Deep Neural Networks (DNNs) have encountered great challenges when involving the applications \cite{krizhevsky2017imagenet,wang2021survey} on resource-constrained devices, owing to the increasing demands for computing and storage resources. Network quantization \cite{lin2016fixed,jacob2018quantization}, which reduces the model size and energy consumption by mapping the floating-point weighs and activations to low-bit ones, is a promising approach to improve the efficiency of DNNs for model compression \cite{han2015deep,qian2021diversifying}. Quantization methods generally dedicate themselves to recovering the \emph{performance drop} originating with the quantization errors, which involves fine-tuning or calibration operations with the original training data.

\begin{figure}[t]
\centering
\setlength{\belowcaptionskip}{-0.35cm}
\includegraphics[width=1.0\columnwidth]{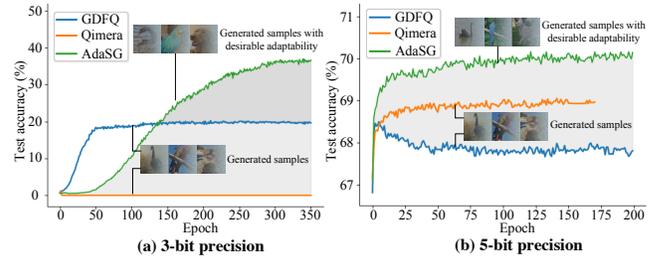}
\caption{Existing work, \emph{e.g.}, GDFQ \cite{xu2020generative} (the \textcolor[RGB]{30,118,180}{blue}) and Qimera \cite{choi2021qimera} (the \textcolor[RGB]{255,127,14}{orange}), suffers from a non-ignorable accuracy loss, as they fail to consider the sample adaptability.
Our AdaSG (the \textcolor[RGB]{44,160,44}{green}) generates the sample with desirable adaptability to maximally recover the performance of Q with varied bit widths, such as (a) 3-bit and (b) 5-bit precision. The observations are from ResNet-18 on ImageNet.  }
\label{drawback}
\end{figure}

However, in many real-world scenes, such as medical and military fields, the original data may not be accessible due to privacy and security issues.
Fortunately, recently proposed data-free quantization (DFQ), a potential method to quantize models without accessing the original data, aims to synthesize meaningful fake samples instead, which improves quantized network (Q) by knowledge distillation \cite{hinton2015distilling,qian2022switchable}  against the pre-trained full-precision model (P).
Among the prior researches \cite{cai2020zeroq,zhang2021diversifying}, the generative fashions \cite{xu2020generative,choi2021qimera,zhu2021autorecon} have recently attracted increasing attention, owing to their superior performance. The generative model is introduced as a generator (G) to capture the distribution of the original data from P for better fake samples, where P is regarded as the discriminator to guide the generation process . For example, Qimera \cite{choi2021qimera} generated boundary supporting samples to reduce the gap between the synthetic and real data.
Nevertheless, \emph{there still remains a non-ignorable performance loss when encountering various bit-width settings}. The reasons may lie in several aspects:
\begin{itemize}
    \item[(1)] Due to the limited capacity of the generator, the generated sample with incomplete distribution is impossible to fully recover the original dataset, so that  it is a crucial criterion: is the sample \emph{beneficial} or \emph{adversarial} over the learning process of Q? However, in the existing arts, the generated sample customized for P, can't always benefit Q in varied bit-width settings (\emph{e.g.}, 3-bit or 5-bit precision, refer to (2)(3)), where only limited information from P can be utilized to recover Q.
    \item[(2)] In 3-bit precision, Q often suffers from a sharp accuracy drop compared to P due to large quantization error, leading to its poor learning ability. Under such case, the generated sample by G may bring an unexpected large \emph{disagreement} between the predictions of P and Q, which makes the optimization loss too large to converge, resulting in no improvement; see the \textcolor[RGB]{255,127,14}{orange} flat curve in Fig.\ref{drawback}(a).
    \item[(3)] In 5-bit precision, Q still possesses comparable recognition ability with P due to a small accuracy drop. Under such case, most of the generated samples by G, for which Q and P give similar predictions (\emph{i.e.}, reach an \emph{agreement}), may not benefit Q. However, under the constraint of the optimization loss, Q receives no improvement, even impairment; see the \textcolor[RGB]{30,118,180}{blue} curve (Epoch 0-50) in Fig.\ref{drawback}(b).
\end{itemize}

\begin{figure}[t]
\setlength{\belowcaptionskip}{-0.35cm}
\centering
\includegraphics[width=1.0\columnwidth]{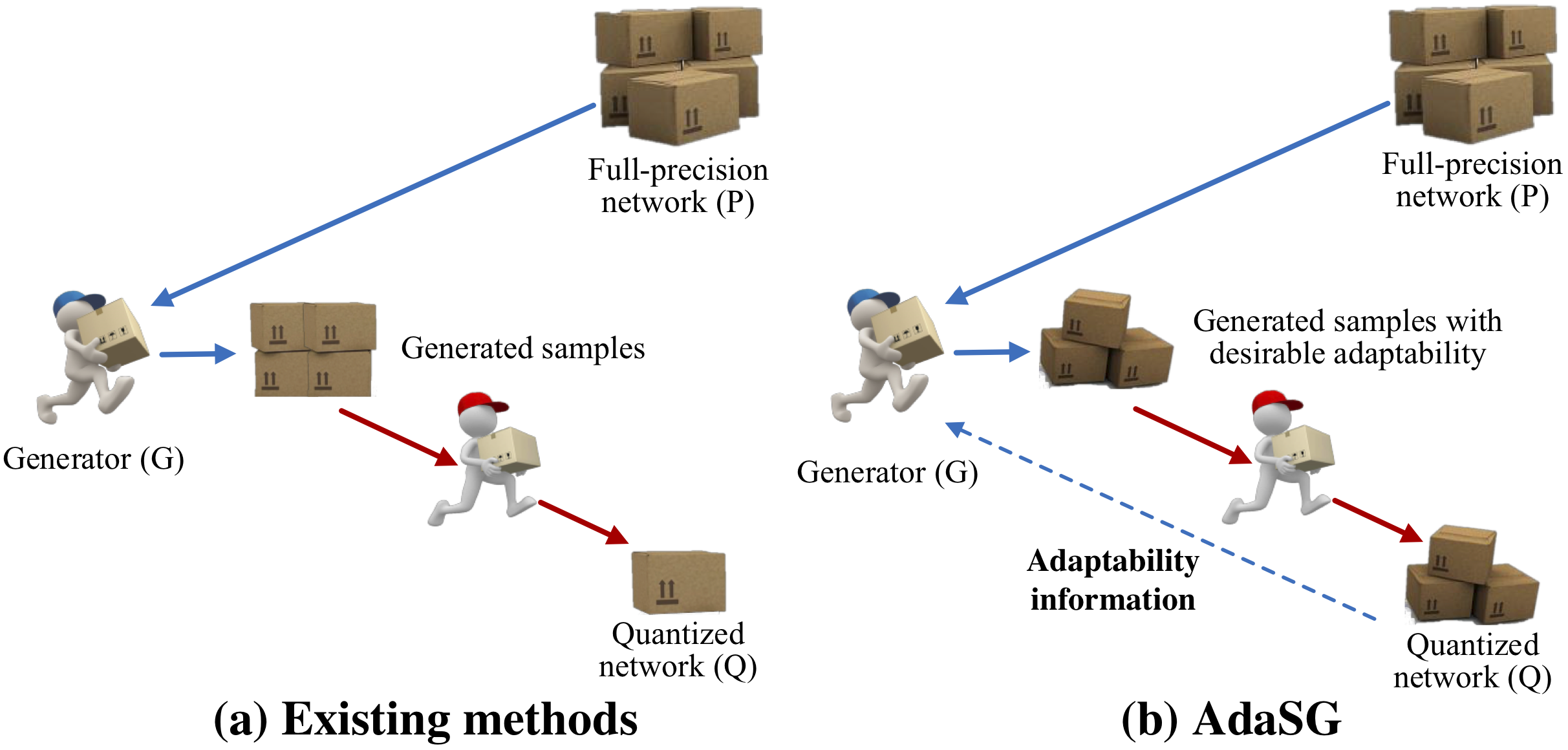}
\caption{Unlike the existing arts (a), our AdaSG (b) aims to generate the sample with desirable adaptability from G, \emph{i.e.}, the dependence of generated sample on Q, to maximally benefit Q with varied bit widths.}
\label{knowledge_medium}
\end{figure}

\noindent Based on the above, the existing approaches fail to consider the \emph{sample adaptability}, \emph{i.e.}, beneficial or adversarial over the calibration process of Q, to Q with varied bit widths during the sample generation process from G, where Q is independent of the generation process; see Fig.\ref{knowledge_medium}(a). For example, for (2), the sample with large adaptability may be one with small disagreement between P and Q; while for (3), it may be one with large disagreement. The above naturally elicits the following basic questions: \emph{how to measure and exploit the sample adaptability to quantized network under varied bit-width scenarios? how to generate the sample with desirable adaptability to benefit the quantized network?}

To answer the above questions, we consider to generate the sample with large adaptability to Q by taking Q into account during the generation process; see Fig.\ref{knowledge_medium}(b).
Specifically, G aims to generate the sample with \emph{large} adaptability, which essentially \emph{maximizes} the reward for G, by enlarging the disagreement between P and Q, to benefit Q; while Q is calibrated to improve itself by exploiting the sample with large adaptability. It is apparent that such sample adaptability will not be large to Q after Q is refined; in other words, such process of benefiting Q leads to \emph{decreasing} the sample adaptability, which essentially \emph{minimizes} the loss for Q, and is adversarial to maximizing the reward goal for G. Based on the above, we rethink date-free quantization process and formulate it as a zero-sum game \cite{li2022policy,zhang2022no} between two players (an adversarial game process where one player's reward is the other's loss while their sum is zero) --- a generator and a quantized network; and we further propose an \textbf{Ada}ptability-aware \textbf{S}ample \textbf{G}eneration (AdaSG) method.
Technically, we specify it via a dynamic maximization-vs-minimization game process anchored on sample adaptability, and defined as:
\begin{equation}
\mathop{min}_{\theta_q\in \Theta_q} \mathop{max}_{\theta_g\in \Theta_g} \  \mathcal{R}(\theta_g,\theta_q),
\label{min_max_s_intro}
\end{equation}
where G and Q are parametrized by $\theta_g\in \Theta_g$ and $\theta_q\in \Theta_q$, respectively.
The optimization of Eq.(\ref{min_max_s_intro}) consists of  fixing $\theta_q$, and updating $\theta_g$ to the optimal $\theta_g^{'}$ for \textit{maximizing } $\mathcal{R}(\theta_g,\theta_q)$; while alternatively fixing $\theta_g^{'}$, and updating $\theta_q$ to the optimal $\theta_q^{'}$ for \textit{minimizing} $\mathcal{R}(\theta_g^{'},\theta_q)$.
Specifically, the maximization process exploits the sample adaptability to Q upon P to generate two types of samples: disagreement (\emph{i.e.}, P can predict correctly but Q not) and agreement samples (\emph{i.e.}, P and Q have the same prediction), such sample adaptability is further reduced by the minimization process after calibrating Q for performance recovery. To achieve the stationarity to maximally benefit Q,
the Balance Gap (BG) for the adaptability of the generated samples within the game process is defined to set up such a stationary objective to balance the maximization versus minimization process.
We remark that AdaSG is essentially an \emph{adversarial} game governed by the sample adaptability, which is fundamentally orthogonal to the existing arts that improve Q by transferring knowledge from P to Q.
One recent study \cite{liu2021zero} generates adversarial samples via G, by maximizing the gap between P and Q, and minimizing their gap to benefit Q for calibration.  However, it fails to consider the sample adaptability to Q, hence suffers from the non-ideal generated samples. Besides, they focus primarily on the adversarial sample generation rather than adversarial game process perspective for AdaSG to generate the samples with desirable adaptability.
The theoretical analysis and empirical studies validate the superiority of AdaSG to the state-of-the-arts.

\section{Adaptability-aware Sample Generation}
Recent generative data-free quantization (DFQ) methods \cite{xu2020generative,choi2021qimera,zhu2021autorecon} aim to reconstruct the original training dataset with a generator (G) by exploiting the distribution from a pre-trained full-precision network (P), which is further exploited to recover the performance of quantized network (Q) via the calibration operation.
However, we observe that Q is independent of the generation process by the existing arts and whether the generated sample is beneficial or adversarial,  namely \emph{sample adaptability}, over the calibration process of Q,  is crucial to the DFQ process.
Motivated by the above, we focus primarily on the sample adaptability to Q. Building on this, the DFQ process is formulated as a dynamic maximization-vs-minimization game process governed by the sample adaptability, as illustrated in Fig.\ref{expand_shrink}. Naturally, one crucial question is how to measure the sample adaptability to Q, which will be discussed in the next section.

\begin{figure}[t]
\centering
\setlength{\belowcaptionskip}{-0.35cm}
\includegraphics[width=0.9\columnwidth]{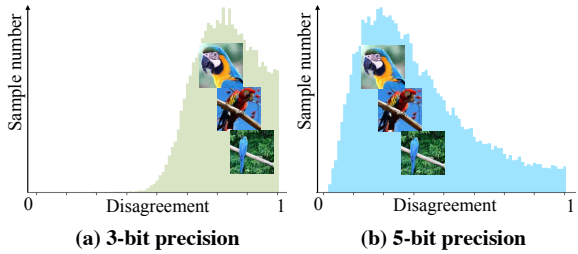}

\caption{Given the 6000 real images from ImageNet as the input to both P and Q upon ResNet-18,  the disagreement between P and Q varies greatly for varied Q with (a) 3-bit and (b) 5-bit precision.
}
\label{dis_diff}
\end{figure}

\subsection{How to Measure the Sample Adaptability to Q?}
To measure the sample adaptability to Q, we primarily focus on two crucial issues: 1) the \emph{dependence} of generated sample on Q, and 2) the \emph{disagreement} between the predictions of P and Q, in that for Q with different bit widths (\emph{e.g.}, 3 or 5 bit), the disagreement varies greatly when delivering the same sample (take real data as an example) to both P and Q; see Fig.\ref{dis_diff}.
We define the generated sample that depends on Q, namely disagreement sample, below:

\noindent \textbf{Definition 1} (Disagreement Sample). \emph{Given a random noise vector $z\sim N(0,1)$ and an arbitrary one-hot label $y$, a generator $G$ generates a sample $x=G(z|y)$. Then the logit outputs of pre-trained full-precision model $P$ and quantized model $Q$ are given as $z_p=P(x)$ and $z_q=Q(x)$, respectively. Suppose that the generated sample $x$ can be correctly predicted by P, i.e., $arg max(z_p)=arg max(y)$. We say $x$ is the disagreement sample if $arg max(z_p)\neq arg max(z_q)$.}

\noindent Thus  the probability vector encoding the disagreement between P and Q, is formulated as
\begin{equation}
\begin{aligned}
&p_{ds}=softmax(z_p- z_q) \in \mathbb{R}^C,
\end{aligned}
\label{dist_disagree}
\end{equation}
where $p_{ds}(c)=\frac{exp(z_p(c)-z_q(c))}{\sum_{j=1}^{C} exp(z_p(j)-z_q(j))}$ ($c\in\{1,2,...,C\}$) represents the $c$-th entry of the vector $p_{ds}$ as the probability that $x$ is labeled as the disagreement sample of the $c$-th class; $C$ denotes the number of class. Together with the sample dependence to Q, the disagreement sample can viewed as the one with large adaptability to Q.

Another key problem is how to model the disagreement between P and Q, which can be exploited to measure the sample adaptability. Eq.(\ref{dist_disagree}) shows that the disagreement reaches the maximum provided $p_{ds}(c)$ approaches to 1 while with other elements to be 0, which corresponds to the minimum entropy of $p_{ds}$; while the disagreement reaches the minimum if each element in $p_{ds}$ is equal, indicating the maximum entropy of $p_{ds}$. Hence, the disagreement can be computed via the information entropy function $\mathcal{H}_{info}(\cdot)$, and formulated as
\begin{small}
\begin{equation}
\mathcal{H}_{info}(p_{ds})=\sum^{C}_{c=1} p_{ds}(c)log\frac{1}{p_{ds}(c)}.
\label{metric}
\end{equation}
\end{small}
For different datasets, $C$ varies greatly, we further normalize $\mathcal{H}_{info}(p_{ds})$ as
\begin{small}
\begin{equation}
\begin{aligned}
\mathcal{H}&=1-\mathcal{H}^{'}_{info}(p_{ds}) \in [0,1),
\end{aligned}
\label{measure_r}
\end{equation}
\end{small}
where $\mathcal{H}^{'}_{info}(p_{ds})=\frac{\mathcal{H}_{info}(p_{ds})-min(\mathcal{H}_{info}(p_{ds}))}{max(\mathcal{H}_{info}(p_{ds}))-min(\mathcal{H}_{info}(p_{ds}))}$. The constant $max(\mathcal{H}_{info}(p_{ds}))=-\sum^{C}_{c=1}\frac{1}{C}log\frac{1}{C}$ represents the maximum value of $\mathcal{H}_{info}(p_{ds})$, in the case that each element in $p_{ds}$ has the same class probability $\frac{1}{C}$, where Q perfectly aligns with P (\emph{i.e.}, $z_p=z_q$); while $min(\cdot)$ is utilized to obtain the minimum of $\mathcal{H}_{info}(p_{ds})$ within a batch. Thus, the sample adaptability is closely related to $\mathcal{H}$ --- the more $\mathcal{H}$, the larger the disagreement over the sample to both P and Q, to yield the samples with large adaptability.

As per Eq.(\ref{dist_disagree})(\ref{measure_r}), $p_{ds}$ distributes in a $C$-dimensional vector space, while $\mathcal{H}$ is a real value.
We hence map that to a $C$-dimensional space $\mathbb{R}^{C}$ while characterizing $\mathcal{H}$, which is achieved via the unit vector $\frac{p_{ds}}{||p_{ds}||}$ ($||\cdot||$ denotes $\ell_2$ norm), then $\mathcal{H}$ is reformulated as
\begin{small}
\begin{equation}
\mathcal{H}_C= \frac{p_{ds}}{||p_{ds}||}\mathcal{H}=\frac{p_{ds}}{||p_{ds}||} (1-\mathcal{H}^{'}_{info}(p_{ds})) \in \mathbb{R}^{C},
\label{h_c}
\end{equation}
\end{small}
where the category information (\emph{i.e.}, $\frac{p_{ds}}{||p_{ds}||}$) of the generated sample apart from the disagreement (\emph{i.e.}, $\mathcal{H}$) is exploited to well measure the sample adaptability.

\begin{figure}[t]
\centering
\setlength{\belowcaptionskip}{-0.35cm}
\includegraphics[width=1.0\columnwidth]{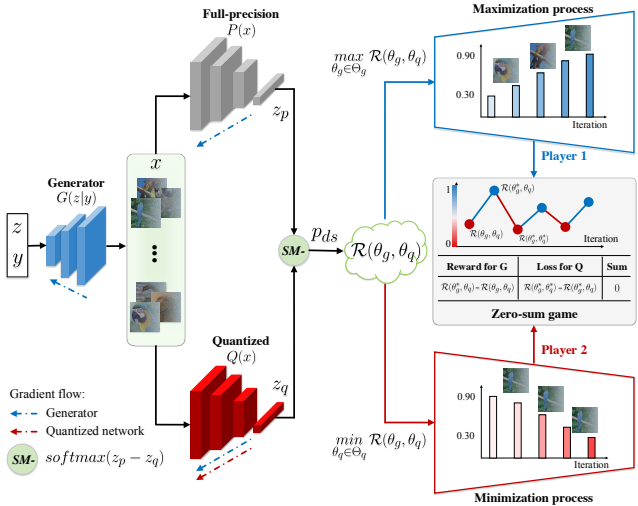}
\caption{Illustration of data-free quantization process as a zero-sum game between generator G (\textcolor[RGB]{0,112,192}{player 1}) and quantized network Q (\textcolor[RGB]{255,0,0}{player 2}). G generates the sample with large adaptability by maximizing $\mathcal{R}(\cdot,\cdot)$ during the maximization process, such sample adaptability is further reduced by minimizing $\mathcal{R}(\cdot,\cdot)$ during the minimization process when calibrating Q.
}
\label{expand_shrink}
\end{figure}

The measurement of sample adaptability inspires us to revisit the DFQ process based on $\mathcal{H}$: the goal of G is to generate the sample with \emph{large} adaptability to Q,  \emph{i.e.}, \emph{increasing} the sample adaptability, by \emph{maximizing} $\mathcal{H}$ (the gain of $\mathcal{H}$ is \emph{positive}, serving as the reward for G),  to benefit Q; while for the calibration process, Q is optimized to recover itself with the generated sample. It infers that such sample adaptability will not be large to Q, while the sample can no longer benefit Q after Q is refined;  in other words, such process of benefiting Q by learning from P results in \emph{decreasing} the sample adaptability, which is achieved by \emph{minimizing} $\mathcal{H}$ (the gain of $\mathcal{H}$ is \emph{negative}, serving as the loss for Q), \emph{adversarial} to maximizing $\mathcal{H}$, which encourages the loss for Q to cancel out the reward for G, such that the sum of reward and loss tends to be zero. Such fact makes the overall changing (summation) of the disagreement ($\mathcal{H}$) \textit{close} to 0 (see Fig.\ref{expand_shrink} for the intuition), which is in line with the principle of zero-sum game  \cite{shoham2008multiagent,v1928theorie},  as discussed in the next section. 


\subsection{Zero-Sum Game: Adversarial Game for Data-Free Quantization}
\label{dfq_game}
We formally revisit the DFQ process from a game theory perspective \cite{li2022policy,zhang2022no}, and formulate the DFQ as a zero-sum game between two players --- a generator and a quantized network, which is an expansion of Eq.(\ref{min_max_s_intro}), as follows:
\begin{small}
\begin{equation}
\mathop{min}_{\theta_q\in \Theta_q} \mathop{max}_{\theta_g\in \Theta_g} \  \mathcal{R}(\theta_g,\theta_q)=\mathop{min}_{\theta_q\in \Theta_q} \mathop{max}_{\theta_g\in \Theta_g} \ \mathbb{E}_{z,y}[1-\mathcal{H}^{'}_{info}(p_{ds})],
\label{max_min}
\end{equation}
\end{small}
where G and Q are parameterized by $\theta_g\in \Theta_g$ and $\theta_q\in \Theta_q$, respectively. In particular, Eq.(\ref{max_min}) is iteratively optimized via gradient descent during the training process, where each iteration consists of two steps: on one hand, $\theta_q$ is fixed, $\mathcal{R}(\theta_g,\theta_q)$ in Eq.(\ref{max_min}) is \emph{maximized} to update $\theta_g$, which is equivalent to generating the sample with \emph{large} adaptability,  \emph{i.e.}, \emph{increasing} the sample adaptability; on the other hand, $\theta_g$ is fixed,  $\mathcal{R}(\theta_g,\theta_q)$ in Eq.(\ref{max_min}) is \emph{minimized} to update $\theta_q$, which is equivalent to calibrating Q with the generated sample, benefiting from \emph{decreasing} the sample adaptability. During the zero-sum game, the overall changing (summation) of the sample adaptability to Q incurred by maximum and minimum optimization is close to zero. 
One critical question is how to converge Eq.(\ref{max_min}) to achieve the equilibrium (\emph{i.e.}, the sample adaptability no longer changes) for the zero-sum game.  Thanks to the classical Nash equilibrium \cite{cardoso2019competing} defined as a state, where no player can improve its individual gain during the zero-sum game, the optimization process will reach an equilibrium $(\theta_g^*,\theta_q^*)$ when (1) for the fixed $\theta_q^*$, G fails to maximize $\mathcal{R}(\theta_g,\theta_q^*)$ to obtain a $\theta_g\in \Theta_g$, yielding $\mathcal{R}(\theta_g,\theta_q^*) > \mathcal{R}(\theta_g^*,\theta_q^*)$; (2) for the fixed $\theta_g^*$, Q can no longer improve itself by minimizing $\mathcal{R}(\theta_g^*,\theta_q)$ to obtain a $\theta_q\in \Theta_q$, yielding $\mathcal{R}(\theta_g^*,\theta_q) < \mathcal{R}(\theta_g^*,\theta_q^*)$. Based on that, for all $\theta_g\in \Theta_g$ and $\theta_q\in \Theta_q$, $(\theta_g^*,\theta_q^*)$ satisfies the following inequality:
\begin{equation}
\mathcal{R}(\theta_g,\theta_q^*) \le \mathcal{R}(\theta_g^*,\theta_q^*) \le \mathcal{R}(\theta_g^*,\theta_q),
\label{}
\end{equation}
where $\theta_g^*$ and $\theta_q^*$ are the parameters of G and Q under an equilibrium state, respectively.

We remark that maximizing $\mathcal{R}(\cdot,\cdot)$ in Eq.(\ref{max_min}) is equivalent to generating the sample with largest adaptability, which may incur too large disagreement (\emph{i.e.}, too small $\mathcal{H}^{'}_{info}(p_{ds})$) between P and Q, while Q (especially for Q with low bit width) has no sufficient ability to learn informative knowledge from P, which, in turn, encourages G to generate the sample with lowest adaptability by alternatively maximizing $\mathcal{R}(\cdot,\cdot)$ in Eq.(\ref{max_min}). However, encouraging the sample with lowest adaptability may incur too large agreement (\emph{i.e.}, too large $\mathcal{H}^{'}_{info}(p_{ds})$) between P and Q, where the generated sample may not be informative to calibrate Q (especially for Q with high bit width). 
The above facts indicate that the sample with either largest or lowest adaptability generated by maximizing $\mathcal{R}(\cdot,\cdot)$ in Eq.(\ref{max_min}) is not necessarily the best.
To address the issues, we further refine the maximization objective of Eq.(\ref{max_min}) during the zero-sum game by the cooperation between the disagreement and agreement sample along with the bound constraints on sample adaptability, which will be elaborated in the next.

%

\begin{figure*}[t]
\centering
\setlength{\belowcaptionskip}{-0.35cm}
\includegraphics[width=0.9\textwidth]{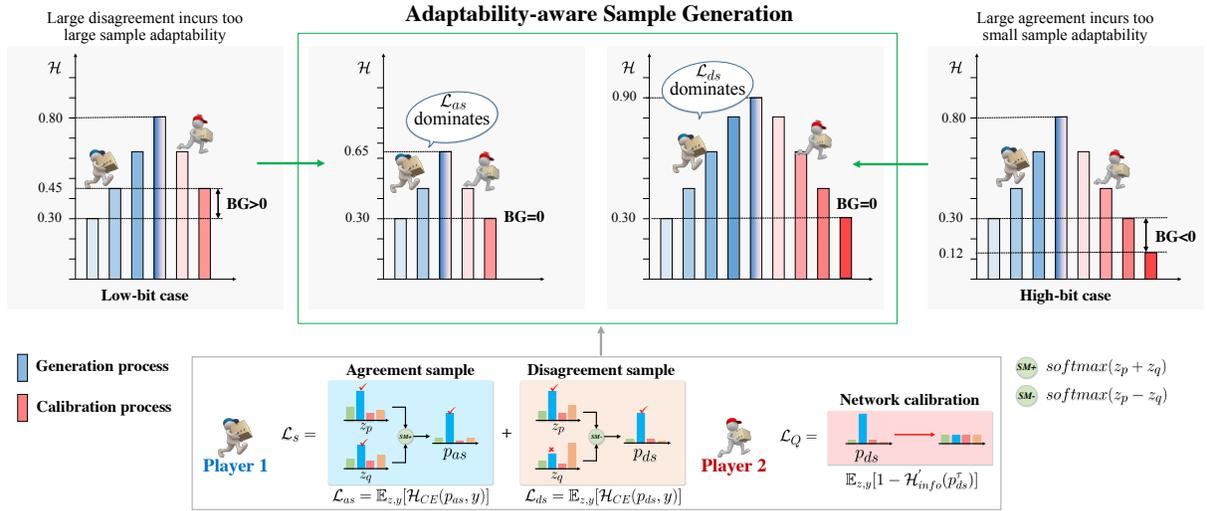}

\caption{Illustration of AdaSG on achieving the stationarity of the zero-sum game via the balance gap (BG). When BG $>0$ (BG $<0$), the generated sample leads to a large disagreement (agreement) between P and Q, owing to the weak (strong) learning ability of Q, \emph{e.g.}, Q with low-bit (high-bit) precision, resulting into too large (small) sample adaptability.  When BG = 0, the DFQ game process is stationary, revealing that the sample adaptability increased by the maximization process can be fully exploited by the minimization process to maximally benefit Q with varied bit widths.
  }
\label{s_ada}
\end{figure*}

\subsection{Maximization Process: Generating Samples with Desirable Adaptability}
\label{expanding_stage}
According to Definition 1, the category (label) information is crucial to establish the dependence of generated sample on Q.  Hence, to generate the disagreement sample with large adaptability and further maximize $\mathcal{R}(\cdot,\cdot)$ in Eq.(\ref{max_min}), we exploit the category (label) information to generate the sample. To achieve it, given the label $y$, the generated sample should be classified as disagreement sample with the same label $y$. Thereby, we present the following Cross-Entropy loss $\mathcal{H}_{CE}(.,.)$ to match $p_{ds}$ and $y$, formulated as
\begin{equation}
\mathcal{L}_{ds}=\mathbb{E}_{z,y}[\mathcal{H}_{CE}(p_{ds},y)].
\label{ds_ce}
\end{equation}
We aim to minimize Eq.(\ref{ds_ce}) to encourage G to generate the disagreement sample that P can predict correctly but Q fails, which, however, may incur too large disagreement (\emph{i.e.}, too small $\mathcal{H}^{'}_{info}(p_{ds})$) between P and Q, thus fail to yield the desirable sample adaptability. To remedy such issue, complementary to the disagreement sample, we further define the agreement sample below:

\noindent \textbf{Definition 2} (Agreement Sample). \emph{ Based on Definition 1, we say the generated sample $x$ is the agreement sample if $arg max(z_p)=arg max(z_q)$.}

\noindent Thus, similar in spirit to $p_{ds}$, the probability vector that describes the agreement between P and Q, is formulated as
\begin{equation}
p_{as}=softmax(z_p+z_q) \in \mathbb{R}^C,
\end{equation}
where $p_{as}(c)=\frac{exp(z_p(c)+z_q(c))}{\sum_{j=1}^{C} exp(z_p(j)+z_q(j))}$ is the $c$-th entry of the vector $p_{as}$ as the probability that $x$ is the agreement sample of the $c$-th class.
Following Eq(\ref{ds_ce}), the loss function for generating agreement sample is given as
\begin{equation}
\mathcal{L}_{as}=\mathbb{E}_{z,y}[\mathcal{H}_{CE}(p_{as},y)],
\label{las}
\end{equation}
which is minimized to encourage to generate the agreement sample that both P and Q can correctly predict to possess larger $\mathcal{H}^{'}_{info}(p_{ds})$.

\subsubsection{Cooperation for desirable sample adaptability}
Based on the above, let $\mathcal{L}_{ds}$ and $\mathcal{L}_{as}$ cooperate with each other to maximize $\mathcal{R}(\cdot,\cdot)$ in Eq.(\ref{max_min}), the sample with desirable adaptability can be generated. Hence, the sample generation loss is given as
\begin{equation}
\mathcal{L}_{s}=\mathcal{L}_{ds} + \mathcal{L}_{as}.
\label{L_G}
\end{equation}
With $\mathcal{L}_{s}$, when a large disagreement incurs, $\mathcal{L}_{as}$ dominates to generate agreement sample, to reduce the gap between P and Q, so as to enlarge $\mathcal{H}^{'}_{info}(p_{ds})$ for disagreement sample; to be analogous, $\mathcal{L}_{ds}$ dominates to reduce $\mathcal{H}^{'}_{info}(p_{ds})$ for agreement sample; see Fig.\ref{s_ada}.
Nevertheless, in some cases, there still exist the samples with either too small or large $\mathcal{H}^{'}_{info}(p_{ds})$ beyond the cooperation ability. 


\subsubsection{Bound constraint on sample adaptability}
\label{boundary_constraint}
To address the above problem, we further impose the bound constraints for $\mathcal{H}^{'}_{info}(p_{ds})$ via the hinge loss \cite{lim2017geometric} below:
\begin{small}
\begin{equation}
\begin{aligned}
\mathcal{L}_b=&\mathbb{E}_{z,y}[max\big(\lambda_l-\mathcal{H}_{info}^{'}(p_{ds}),0\big)] \\
&+ \mathbb{E}_{z,y}[max\big(\mathcal{H}_{info}^{'}(p_{ds})-\lambda_u,0\big)],
\label{boundary_loss}
\end{aligned}
\end{equation}
\end{small}
where $\lambda_l$ and $\lambda_u$ denote the lower and upper bound of $\mathcal{H}^{'}_{info}(p_{ds})$, such that $0 \le \lambda_l < \lambda_u \le 1$. 
Specifically, $\lambda_l$ serves to prevent G from generating the sample with too large adaptability (\emph{i.e.}, too small $\mathcal{H}^{'}_{info}(p_{ds})$) via maximizing Eq.(\ref{max_min}), while $\lambda_u$ aims to avoid the sample with too small adaptability (\emph{i.e.}, too large $\mathcal{H}^{'}_{info}(p_{ds})$), which, in turn, offer the guarantee for the cooperation between $\mathcal{L}_{ds}$ and $\mathcal{L}_{as}$. 
\subsubsection{BNS information}
The above fact discusses the adaptability of single sample to Q, while calibrating Q requires the generated samples within a batch, motivating us to exploit the distribution information of the training data.
We consider the batch normalization statistics (BNS) information \cite{xu2020generative,choi2021qimera} about the training data contained in P, which is learned by
\begin{small}
\begin{equation}
\mathcal{L}_{BNS}=\sum^M_{m=1}( ||\mu^g_m-\mu_m||^2_2 + ||\sigma^g_m-\sigma_m||^2_2 ),
\label{bns}
\end{equation}
\end{small}
where $\mu^g_m / \sigma^g_m$ are the mean/variance of the generated sample's distribution at the $m$-th BN layer of the total $M$ layers; and $\mu_m / \sigma_m$ are the corresponding mean/variance parameters stored in the $m$-th BN layer of P.

\subsubsection{Overall loss}
To this end, we finalize the loss function for the maximization (sample generation) process as
\begin{equation}
\mathcal{L}_G=  \alpha (\mathcal{L}_{ds}+ \mathcal{L}_{as}) + \beta \mathcal{L}_b+ \gamma \mathcal{L}_{BNS},
\label{generator_loss}
\end{equation}
where $\alpha$, $\beta$ and $\gamma$ are the balance hyperparameters.
Throughout minimizing $\mathcal{L}_G$, which is equivalent to maximizing $\mathcal{R}(\cdot,\cdot)$ in Eq.(\ref{max_min}), the sample with desirable adaptability can be generated, such sample is exploited by calibrating Q during the minimization process according to Eq.(\ref{max_min}).

\subsection{Minimization Process: Calibrating Q by Reducing Sample Adaptability}
\label{shrinking_stage}
With the above generated sample, the goal of the minimization process is to calibrate Q for the performance recovery by minimizing $\mathcal{R}(\cdot,\cdot)$ in Eq.(\ref{max_min}), which indicates that the disagreement between P and Q is decreased to reach the agreement.
In particular, we further introduce a temperature parameter $\tau$ to soften the output, thus the loss function for the minimization (calibration) process is formulated as 
\begin{small}
\begin{equation}
\mathcal{L}_Q=\mathbb{E}_{z,y}[1-\mathcal{H}_{info}^{'}(p_{ds}^{\tau})],
\label{calibration_loss}
\end{equation}
\end{small}
where $p_{ds}^{\tau}=softmax((z_p-z_q)/\tau)$. 
By alternatively optimizing Eq.(\ref{generator_loss}) and (\ref{calibration_loss}) during a zero-sum game, the samples with desirable adaptability can be generated by G to maximally recover the performance of Q until reaching a Nash equilibrium.

\subsection{Theoretical Analysis: a Balance Gap for Sample Adaptability}
One may wonder whether the above maximization-vs-minimization process can keep stationary during the optimization process over Eq.(\ref{max_min}), which is critical for sample generation with desirable adaptability and quantized network recovery.  We confirm that by our theoretical analysis based on the balance gap (BG) over sample adaptability, which is defined as:

\noindent \textbf{Definition 3} (Balance Gap). \emph{Considering the objective $\mathcal{R}(\cdot,\cdot)$ for the DFQ game process. Assume that, after one iteration, the  parameters $(\theta_g^1,\theta_q^1)$ of the game are updated to $(\theta^2_g,\theta_q^2)$ via the gradient descent. Then, the balance gap (BG) is defined as}
\begin{equation}
\text{\emph{BG}} =\mathcal{R}(\theta^2_g,\theta^2_q)-\mathcal{R}(\theta_g^1,\theta_q^1),
\label{}
\end{equation}
which encodes the deviation for the value of $\mathcal{R}(\cdot,\cdot)$ before and after each iteration. When BG $>0$ or BG $<0$, the generated sample leads to a large disagreement or agreement  between P and Q, owing to the weak or strong learning ability of Q, where the  sample with too large or small adaptability fails to benefit Q during the calibration process.
We remark that BG is well bounded to avoid the large deviation, which facilitates generating the samples with desirable adaptability.
Such fact is validated via the following proposition:

\noindent \textbf{Proposition 1.} \emph{Considering the stationarity of DFQ game process. Then, the balance gap (BG) is bounded below: }
\begin{small}
\begin{equation}
\begin{aligned}
|\text{\emph{BG}}|=|\mathcal{R}(\theta^2_g,\theta^2_q)-\mathcal{R}(\theta_g^1,\theta_q^1)| \le L||[\theta^2_g;\theta^2_q]-[\theta_g^1;\theta_q^1]||
\end{aligned}
\label{bg_upper}
\end{equation}
\end{small}
\emph{where $L$ is the Lipschitz constant.}

\noindent \emph{\textbf{Proof.} }
\emph{Considering the Lipschitz continuity of the objective function $\mathcal{R}(\cdot,\cdot)$, we divide the proof into two parts: }

\noindent \emph{\textbf{Part 1.} Lipschitz Continuity}

\noindent \emph{We say a function f(x,y) is L-Lipschitz continuous with respect to a norm $||.||$ if }
\begin{small}
\begin{equation}
|f(x_2,y_2)-f(x_1,y_1)| \le L||[x_2;y_2] - [x_1;y_1]||
\label{}
\end{equation}
\end{small}
\emph{for any $x_1,x_2\in X$ and any $y_1,y_2\in Y$. The previous inequality holds if and
only if}
\begin{small}
\begin{equation}
||[\nabla_x f(x,y);\nabla_y f(x,y)]|| \le L
\label{}
\end{equation}
\end{small}
\emph{for any $x\in X$ and any $y\in Y$. L is the Lipschitz constant.}

\noindent \emph{\textbf{Part 2.}}
\noindent \emph{According to the \textbf{first-order Taylor expansion} and \textbf{Cauchy inequality}, we have:}
\begin{equation}
\begin{small}
\begin{aligned}
|\text{\emph{BG}}|&=|\mathcal{R}(\theta^2_g,\theta^2_q)-\mathcal{R}(\theta_g^1,\theta_q^1)| \\
&= || [\nabla_{\theta_g} \mathcal{R}(\theta_g,\theta_q);\nabla_{\theta_q} \mathcal{R}(\theta_g,\theta_q)]^T \  [\theta^2_g-\theta_g^1;\theta^2_q-\theta_q^1] || \\
&\le || [\nabla_{\theta_g} \mathcal{R}(\theta_g,\theta_q);\nabla_{\theta_q} \mathcal{R}(\theta_g,\theta_q)]^T|| \  ||[\theta^2_g;\theta^2_q]-[\theta_g^1;\theta_q^1]||,
\end{aligned}
\end{small}
\label{}
\end{equation}
 \emph{then Eq.(\ref{bg_upper}) holds if and only if $\nabla_{\theta_g} \mathcal{R}(\theta_g,\theta_q)$ and $\nabla_{\theta_q} \mathcal{R}(\theta_g,\theta_q)$ are upper bounded. }
\noindent \emph{First, we note that }
\begin{equation}
\begin{small}
\begin{aligned}
\mathcal{R}(\theta_g,\theta_q)=\mathbb{E}_{z,y}[1-\mathcal{H}^{'}_{info}(p_{ds})],
\end{aligned}
\end{small}
\label{}
\end{equation}
\noindent \emph{where $\mathcal{H}^{'}_{info}(p_{ds})$ is obtained by normalizing $\mathcal{H}_{info}(p_{ds})=\sum^{C}_{c=1} p_{ds}(c)log\frac{1}{p_{ds}(c)}$, and $p_{ds}=softmax(z_p- z_q)$; }
\noindent \emph{and the corresponding gradients are computed by }
\begin{equation}
\begin{small}
\begin{aligned}
& \nabla_{\theta_g} \mathcal{R}(\theta_g,\theta_q)=\frac{\partial \mathcal{R}(\theta_g,\theta_q)}{\partial p_{ds}}   \frac{\partial p_{ds}}{\partial x} \frac{\partial x}{\partial \theta_g},  \\
& \nabla_{\theta_q} \mathcal{R}(\theta_g,\theta_q)=\frac{\partial \mathcal{R}(\theta_g,\theta_q)}{\partial p_{ds}}   \frac{\partial p_{ds}}{\partial z_q} \frac{\partial z_q}{\partial \theta_q},
\end{aligned}
\end{small}
\label{}
\end{equation}
\noindent \emph{where $\frac{\partial x}{\partial \theta_g}$ and $\frac{\partial z_q}{\partial \theta_q}$ are the gradients produced by the generator and quantized network, respectively. }
\emph{It is apparent that the gradients of the information entropy function $\mathcal{H}_{info}(\cdot)$, the softmax function and the activation function ({e.g.}, Sigmod, Relu) in deep neural network are \textbf{all bounded}. Therefore, $\forall L \in \mathbb{R}$, make $|| [\nabla_{\theta_g} \mathcal{R}(\theta_g,\theta_q);\nabla_{\theta_q} \mathcal{R}(\theta_g,\theta_q)]^T|| $ be upper bounded.   }
\noindent \emph{To sum up, the balance gap (BG) is bounded by Eq.(\ref{bg_upper}). }

 \hfill $\square$

The proposition provides an upper and lower bound for BG, to well avoid the sample generation encoding too large or small adaptability, in line with the intuition of $\mathcal{L}_b$, which offers a guarantee for the stationarity of DFQ game, we further achieve the stationarity via the following proposition:

\noindent \textbf{Proposition 2.} \emph{During each iteration, the DFQ game is stationary, that is, the sample adaptability increased by the maximization process can be fully exploited by the minimization process to maximally benefit Q, when BG = 0. }

\noindent \emph{\textbf{Proof.} }
\noindent \emph{As aforementioned, each iteration contains the maximization and minimization process, then we have:}
\noindent \emph{
\begin{equation}
\begin{small}
\begin{aligned}
\text{{BG}}&=\mathcal{R}(\theta^2_g,\theta^2_q)-\mathcal{R}(\theta_g^1,\theta_q^1) \\
&=\big(\mathcal{R}(\theta^2_g,\theta_q^1) - \mathcal{R}(\theta_g^1,\theta_q^1)\big) - \big(\mathcal{R}(\theta^2_g,\theta_q^1) - \mathcal{R}(\theta^2_g,\theta^2_q)\big)  \\
&=\Delta_g-\Delta_q,
\end{aligned}
\end{small}
\label{}
\end{equation}
where $\Delta_g$ and $\Delta_q$ denote the \textbf{deviation} for the value of $\mathcal{R}(.,.)$ during the \textbf{maximization} and \textbf{minimization} process, respectively.
}
 \emph{
Thus, if \bm{$\Delta_g >\Delta_q$} (i.e., $\text{{BG}}>0$), the generated sample leads to a large disagreement between P and Q, owing to the weak learning ability of Q; if \bm{$\Delta_g <\Delta_q$} (i.e., $\text{{BG}}<0$), the generated sample leads to a large agreement between P and Q, owing to the strong learning ability of Q. Under such cases, the sample with too large or small adaptability fails to benefit Q during the calibration process. 
}
\noindent \emph{
To sum up, if and only if \bm{$\Delta_g =\Delta_q$} (i.e., $\text{{BG}}=0$), the DFQ game is stationary, that is, the sample adaptability increased by the maximization process can be fully exploited by the minimization process to maximally benefit Q by minimizing $L_Q$, which is equivalent to minimizing  $\mathcal{R}(\cdot,\cdot)$  in Eq.(\ref{max_min}).
}

\hfill $\square$

The proposition discloses the effectiveness of the coordination between $\mathcal{L}_{ds}$ and $\mathcal{L}_{as}$ during the maximization process. Specifically, when BG $>0$ ($< 0$), the sample adaptability is too large or small; therefore, for next iteration, $\mathcal{L}_{as}$ ($\mathcal{L}_{ds}$) will dominate to generate the sample with desirable adaptability by minimizing $L_G$ (equivalent to maximizing $\mathcal{R}(\cdot,\cdot)$  in Eq.(\ref{max_min})), to achieve the stationarity of the DFQ game, \emph{i.e.}, BG $=0$; see Fig.\ref{s_ada}.


\section{Experiment}
\subsection{Experimental Settings and Implementation Details}
We validate AdaSG over three typical image classification datasets, including \textbf{CIFAR-10}, \textbf{CIFAR-100} \cite{krizhevsky2009learning} and \textbf{ImageNet} (\textbf{ILSVRC2012}) \cite{russakovsky2015imagenet}. CIFAR-10 and CIFAR-100 contain 10 and 100 classes of images, respectively. Both of them are split into 50K training images and 10K testing images. ImageNet consists of 1.2M samples for training and 50k samples for validation with 1000 categories. For data-free setting, only validation sets are adopted to evaluate the performance of the quantized models.
In the experiments, we quantize pre-trained full-precision networks P including ResNet-20 for CIFAR, and ResNet-18, ResNet-50, and MobileNetV2 for ImageNet, via the following quantizer to yield the quantized networks Q:

\noindent \textbf{Quantizer.}
Following \cite{xu2020generative,choi2021qimera}, we quantize both full-precision (float32) weights and activations into $n$-bit precision via a symmetric linear quantization method based on \cite{jacob2018quantization} below:
\begin{small}
\begin{equation}
\theta_q=round \Big ((2^n-1) * \frac{\theta-\theta_{min}}{\theta_{max}-\theta_{min}}-2^{n-1} \Big),
\end{equation}
\end{small}
where $\theta$ and $\theta_q$ are the full-precision and quantized value. $round(\cdot)$ returns the nearest integer value to the input. $\theta_{min}$ and $\theta_{max}$ are the minimum and maximum of $\theta$, respectively.
Regarding the bit width $n$, we select 3-bit, 4-bit and 5-bit precision, which are representative for low-bit and high-bit cases, particularly: 3-bit quantization actually leads to a huge performance loss, which is a major challenge for the existing DFQ methods; while 5-bit or higher-bit quantization usually causes a small performance loss, which is selected to validate the generalization ability. 

For the \emph{maximization} process, we construct the architecture of the generator (G) following ACGAN \cite{odena2017conditional}, while P and Q play the role of discriminator, where G is trained with the loss function Eq.(\ref{generator_loss}) using Adam \cite{kingma2014adam} as an optimizer with a momentum of 0.9 and a learning rate of 1e-3.
For the \emph{minimization} process, Q is optimized with the loss function Eq. (\ref{calibration_loss}), where SGD with Nesterov \cite{nesterov1983method} is adopted as an optimizer with a momentum of 0.9 and weight decay of 1e-4. For CIFAR, the learning rate is initialized to 1e-4 and decayed by 0.1 for every 100 epochs, while it is 1e-5 (1e-4 for ResNet-50) and divided by 10 at epoch 350 (at epoch 200 and 300 for ResNet-50) on ImageNet. The generator and quantized model are totally trained for 400 epochs. The batch size is set to 16.
For the hyperparameters,  $\alpha$, $\beta$ and $\gamma$ in Eq.(\ref{generator_loss}); $\lambda_l$ and $\lambda_u$ in Eq.(\ref{boundary_loss}) are empirically set to be 0.1, 1, 1, 0.3 and 0.8. 
All experiments are implemented with pytorch \cite{paszke2019pytorch} based on the code of GDFQ \cite{xu2020generative} and run on an NVIDIA GeForce GTX 1080 Ti GPU and an Intel(R) Core(TM) i7-6950X CPU @ 3.00GHz.

To validate how AdaSG generates the sample with desirable adaptability to maximally benefit Q, we empirically validate why AdaSG works, including the comparisons with the  state-of-the-arts, ablation study as well as visual analysis.

\begin{table*}
\setlength{\abovecaptionskip}{0.02cm}
\caption{Accuracy (\%) comparison with the state-of-the-arts on CIFAR-10, CIFAR-100 and ImageNet. $\dag$: the results implemented by author-provided code. -: no results are reported. \emph{n}w\emph{n}a indicates the weights and activations are quantized to \emph{n} bit. The best results are reported with \textbf{boldface}.}
\label{sota}
\centering
\scriptsize
\begin{tabular}{c|cc|c|  c| c| ccc  |c}
\toprule
\textbf{Dataset} &\tabincell{c}{\textbf{Model} \\ (Full precision)}     &\textbf{Bit width}  &\tabincell{c}{\textbf{ZAQ} \\ (CVPR 2021)} &\tabincell{c}{\textbf{IntraQ}  \\ (CVPR 2022)} &\tabincell{c}{\textbf{ARC+AIT}  \\ (CVPR 2022)}  &\tabincell{c}{\textbf{GDFQ} \\ (ECCV 2020)}    &\tabincell{c}{\textbf{ARC} \\ (IJCAI 2021)}    &\tabincell{c}{\textbf{Qimera} \\ (NeurIPS 2021)}    &\tabincell{c}{\textbf{AdaSG} \\ \textbf{(Ours)}}   \\
\bottomrule
\toprule
\multirow{3}{*}{\textbf{CIFAR-10}}
&                       &3w3a     &-      &77.07 &-  &\ 75.11$^{\dag}$     &-     &\ 74.43$^{\dag}$            &\textbf{84.14}   \\
&\emph{ResNet-20}      &\cellcolor{mygreen}4w4a     &\cellcolor{mygreen}92.13 &\cellcolor{mygreen}91.49 &\cellcolor{mygreen}90.49     &\cellcolor{mygreen}90.11     &\cellcolor{mygreen}88.55     &\cellcolor{mygreen}91.26    &\cellcolor{mygreen}\textbf{92.10}  \\
&(93.89)            &5w5a     &93.36 &-  &92.98   &93.38     &92.88     &93.46   &\textbf{93.76}   \\
\bottomrule
\toprule
\multirow{3}{*}{\textbf{CIFAR-100}}
&                       &3w3a     &-    &48.25 &-  &\ 47.61$^{\dag}$      &-    &\ 46.13$^{\dag}$           &\textbf{52.76}   \\
&\emph{ResNet-20}      &\cellcolor{mygreen}4w4a     &\cellcolor{mygreen}60.42 &\cellcolor{mygreen}64.98 &\cellcolor{mygreen}61.05    &\cellcolor{mygreen}63.75      &\cellcolor{mygreen}62.76    &\cellcolor{mygreen}65.10   &\cellcolor{mygreen}\textbf{66.42}  \\
&(70.33)            &5w5a     &68.70 &-  &68.40   &67.52      &68.40    &69.02   &\textbf{69.42}   \\

\bottomrule
\toprule
\multirow{3}{*}{\textbf{}}
&                       &3w3a    &-      &- &-  &\ 20.23$^{\dag}$   &23.37       &\ 1.17$^{\dag}$    &\textbf{37.04}   \\
&\emph{ResNet-18}      &\cellcolor{mygreen}4w4a    &\cellcolor{mygreen}52.64 &\cellcolor{mygreen}66.47 &\cellcolor{mygreen}65.73     &\cellcolor{mygreen}60.60   &\cellcolor{mygreen}61.32       &\cellcolor{mygreen}63.84    &\cellcolor{mygreen}\textbf{66.50}  \\
&(71.47)            &5w5a    &64.54 &69.94  &70.28    &68.49   &68.88       &69.29    &\textbf{70.29}   \\

\cline{2-10}
\multirow{3}{*}{\textbf{ImageNet}}
&                       &3w3a    &-     &- &- &\ 1.46$^{\dag}$     &14.30    &-            &\textbf{26.90}   \\
&\emph{MobileNetV2}  &\cellcolor{mygreen}4w4a    &\cellcolor{mygreen}0.10  &\cellcolor{mygreen}65.10 &\cellcolor{mygreen}\textbf{66.47}   &\cellcolor{mygreen}59.43    &\cellcolor{mygreen}60.13    &\cellcolor{mygreen}61.62    &\cellcolor{mygreen}65.15  \\
&(73.03)           &5w5a    &62.35 &71.28 &\textbf{71.96}    &68.11   &68.40   &70.45    &{71.61}   \\

\cline{2-10}
\multirow{3}{*}{\textbf{}}
&                       &3w3a      &-    &-  &- &\ 0.31$^{\dag}$      &1.63    &-            &\textbf{16.98}   \\
&\emph{ResNet-50}      &\cellcolor{mygreen}4w4a      &\cellcolor{mygreen}53.02  &\cellcolor{mygreen}-  &\cellcolor{mygreen}68.27 &\cellcolor{mygreen}54.16    &\cellcolor{mygreen}64.37   &\cellcolor{mygreen}66.25    &\cellcolor{mygreen}\textbf{68.58}  \\
&(77.73)            &5w5a      &73.38  &-  &76.00  &71.63    &74.13   &75.32    &\textbf{76.03}   \\
\bottomrule
\end{tabular}
\vspace{-0.8em}
\end{table*}

\subsection{Why does AdaSG Work?}
\begin{figure}[t]
\centering
\setlength{\abovecaptionskip}{0.02cm}
\setlength{\belowcaptionskip}{-0.35cm}
\includegraphics[width=1.0\columnwidth]{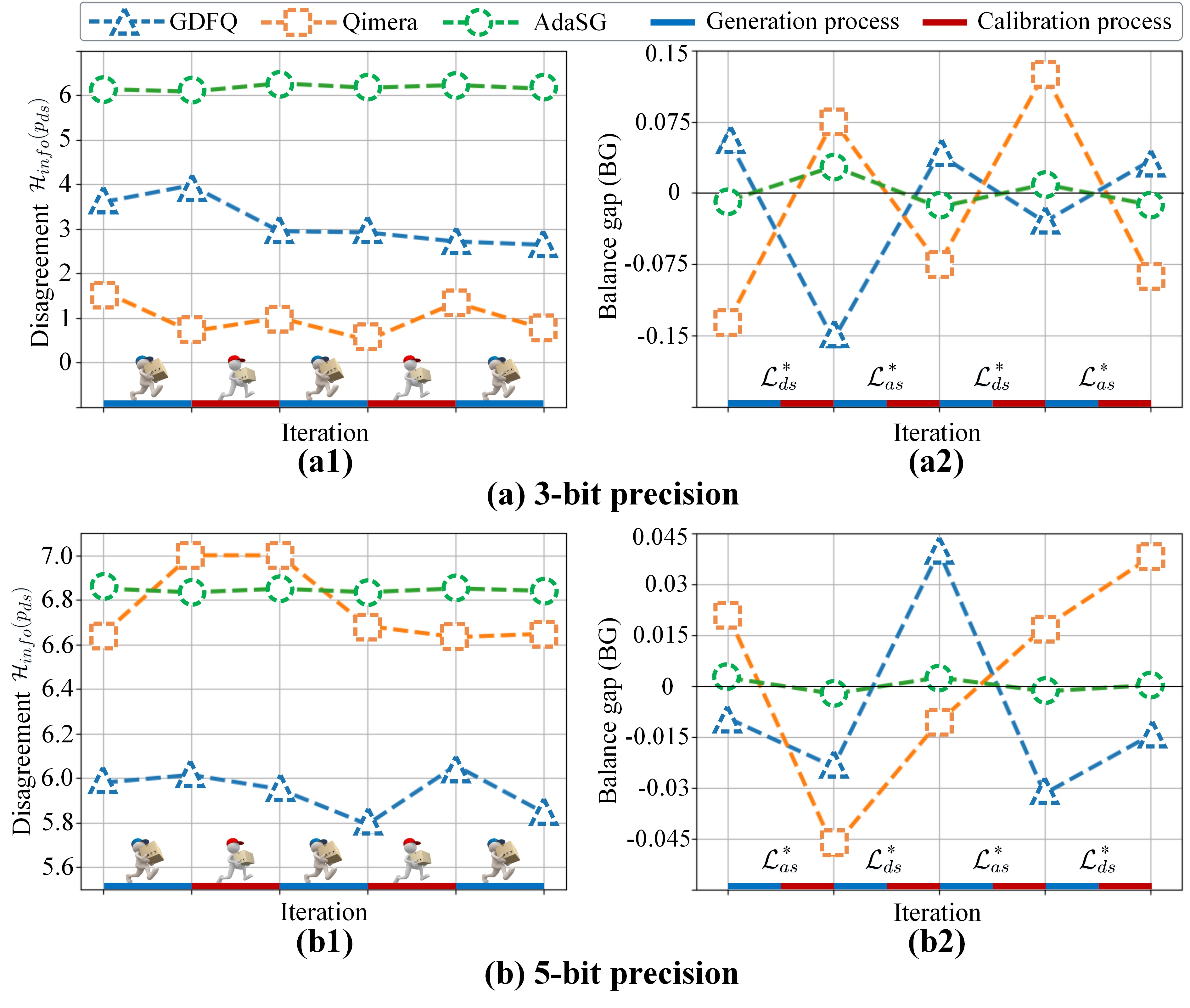}
\caption{Illustration of AdaSG on generating the samples with desirable adaptability to Q under (a) 3-bit and (b) 5-bit precision. {(a1)(b1)} Disagreement (\emph{i.e.}, $\mathcal{H}_{info}(p_{ds})$) between P and Q during the generation (\textcolor[RGB]{0,112,192}{\textbf{---}}) and calibration (\textcolor[RGB]{192,0,0}{\textbf{---}}) process. (a2)(b2) Balance gap (BG).}
\label{ada_experiment}
\end{figure}

We experimentally verify the core idea of AdaSG --- generating the sample with desirable adaptability to recover the performance of Q with varied bit widths. We perform the data-free quantization experiments with ResNet-18 (3-bit and 5-bit precision) serving as both P and Q on ImageNet. 
Fig.\ref{ada_experiment}(a1)(b1) illustrates that the disagreement ($\mathcal{H}_{info}(p_{ds})$ in Eq.(\ref{metric})) between P and Q for AdaSG fluctuates stably within a small range compared to GDFQ \cite{xu2020generative} and Qimera \cite{choi2021qimera}, confirming that the generated sample with desirable adaptability by AdaSG is fully exploited to benefit Q, and the bound constraint loss (Eq.(\ref{boundary_loss})) can avoid generating the sample with too large or small adaptability, which leads to ridiculously large disagreement or agreement.
Fig.\ref{ada_experiment}(a2)(b2) reveals that the balance gap (Definition 3) keeps on par with zero (either BG $>0$ or BG $<0$) during the iteration process unlike GDFQ and Qimera, which confirm the principle that AdaSG achieves the stationarity of the DFQ game and the coordination between $\mathcal{L}_{ds}$ (Eq.(\ref{ds_ce})) and $\mathcal{L}_{as} $ (Eq.(\ref{las})) can generate the sample with desirable adaptability to maximally benefit Q.

\begin{table}[t]
\setlength{\abovecaptionskip}{0.12cm}
\setlength{\belowcaptionskip}{-0.25cm}
\caption{Ablation study about varied components of AdaSG. \emph{n}w\emph{n}a indicates the weights and activations are quantized to \emph{n} bit. The best results are reported with \textbf{boldface}.}
\centering
\scriptsize
\begin{tabular}{c|c|c|c|c|c|c}
\toprule

\multirow{2}{*}{\tabincell{c}{\textbf{Model} \\ (Full precision)}} & \multirow{2}{*}{$\mathcal{L}_{ds}$} & \multirow{2}{*}{$\mathcal{L}_{as}$} & \multirow{2}{*}{$\mathcal{L}_b$} & \multirow{2}{*}{$\mathcal{L}_{BNS}$}  &\multicolumn{2}{c}{\textbf{Bit width}}  \\
\cline{6-7}
&&&&&3w3a &5w5a  \\

\bottomrule
\toprule
\multirow{6}{*}{\tabincell{c}{\emph{ResNet-18} \\ (71.47)}}
&&\ding{51}      &\ding{51}           &\ding{51}                  &14.86     &69.15   \\
&\cellcolor{mygreen}\ding{51}&\cellcolor{mygreen}      &\cellcolor{mygreen}\ding{51}           &\cellcolor{mygreen}\ding{51}                  &\cellcolor{mygreen}26.41     &\cellcolor{mygreen}69.81  \\
&\ding{51}&\ding{51}      &           &\ding{51}                  &32.75     &70.13  \\
&\cellcolor{mygreen}&\cellcolor{mygreen}      &\cellcolor{mygreen}\ding{51}           &\cellcolor{mygreen}\ding{51}                                &\cellcolor{mygreen}15.01     &\cellcolor{mygreen}70.06   \\
&\ding{51}&\ding{51}      &\ding{51}           &                  &20.98     &67.35  \\
&\cellcolor{mygreen}\ding{51} &\cellcolor{mygreen}\ding{51}      &\cellcolor{mygreen}\ding{51}           &\cellcolor{mygreen}\ding{51}    &\cellcolor{mygreen}\textbf{37.04}     &\cellcolor{mygreen}\textbf{70.29}  \\

\bottomrule
\end{tabular}
\label{ablation}
\vspace{-2.4em}
\end{table}

\subsection{Comparison with State-of-the-arts}
To verify the superiority of AdaSG, we compare it with typical DFQ approaches, including: GDFQ \cite{xu2020generative}, ARC \cite{zhu2021autorecon} and Qimera \cite{choi2021qimera}: reconstructing the original data from P; ZAQ \cite{liu2021zero}  focuses primarily on the adversarial sample generation rather than adversarial game process for AdaSG; IntraQ \cite{zhong2022intraq} optimizes the noise to obtain fake sample without a generator; AIT \cite{choi2022s} focuses on improving the loss function and manipulating the gradients for ARC to generate better sample, denoted as ARC+AIT.

Table \ref{sota} summarizes our following observations: 1) AdaSG offers a significant and consistent performance gain over the state-of-the-arts, in line with our purpose of generating the sample with desirable adaptability to maximally benefit Q. Impressively, AdaSG achieves at most 9.71\%, 6.63\% and 35.87\% accuracy gains on CIFAR-10, CIFAR-100 and ImageNet, respectively. Especially, compared with GDFQ, ARC and Qimera where Q is independent of the generation process, AdaSG obtains accuracy improvement with a large margin, \emph{e.g.}, at least 0.3\% gain (ResNet-20 with 5w5a on CIFAR-10), confirming the necessity of AdaSG focusing on the sample adaptability to Q.   Specifically, ZAQ suffers from a large performance gap compared to AdaSG, since many unexpected samples are generated without considering the sample adaptability, which is harmful to calibrating Q.  AdaSG shows obvious advantages over AIT despite of the combination with ARC.
2) AdaSG achieves the substantial gains for Q with varied bit-widths, confirming the desirable adaptability of our generated sample to varied Q. Note that, for 3-bit case, most of the existing methods suffer from a poor accuracy, even fail to converge, while AdaSG obtains at most 35.87\% (ResNet-18 with 3w3a) and at least 4.21\% (ResNet-20 with 3w3a on CIFAR-100) performance gains.

\subsection{Ablation Study}

\textbf{Validating adaptability with disagreement and agreement samples}
As aforementioned, the disagreement and agreement samples play a critical role on the sample adaptability to Q, which serve as a pivotal role between the maximization and minimization process during the zero-sum game. We conduct the ablation study on $\mathcal{L}_{ds}$ (Eq.(\ref{ds_ce})) and $\mathcal{L}_{as} $ (Eq.(\ref{las})) over ImageNet. Table \ref{ablation} suggests the great superiority (37.04\% and 70.29\%) of AdaSG (including the both) to other cases. It is worth noting that, removing either or both of $\mathcal{L}_{ds}$ and $\mathcal{L}_{as} $ obtains a large performance degradation (at most 22.18\% and 1.14\%), implying the intuition of the cooperation between $\mathcal{L}_{ds}$ and $\mathcal{L}_{as} $.
Interestingly, the case without $\mathcal{L}_b$ (Eq.(\ref{boundary_loss})) receives the minimal accuracy loss (4.29\% and 0.16\%), confirming the importance of $\mathcal{L}_b$ on the basis of $\mathcal{L}_{ds}$ and $\mathcal{L}_{as}$.

\subsection{Visualization of Generated Samples}
\begin{figure}[t]
\centering
\setlength{\abovecaptionskip}{0.12cm}
\setlength{\belowcaptionskip}{-0.35cm}
\includegraphics[width=1.0\columnwidth]{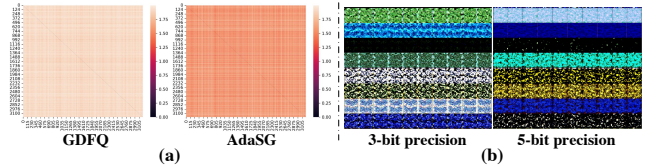}
\caption{{(a)} The similarity comparison between the generated samples. {(b)} Visualization of the generated samples, where each row denotes one of 8 randomly chosen classes from ImageNet.}
\label{simi-visual}
\end{figure}

To further show the desirable adaptability of the generated samples by AdaSG to Q, we conduct the visual analysis over MobileNetV2 serving as both P and Q on ImageNet by the similarity matrix (each element is obtained by computing the $\ell_1$ norm between the probability distribution $p_{ds}$ of every two samples), along with the visualization of generated samples in Fig.\ref{simi-visual}. Fig.\ref{simi-visual}(a) illustrates that the generated samples by AdaSG possess a much larger similarity (the darker, the larger) than those by GDFQ, implying that the generated sample by GDFQ varies greatly, where a lots of samples with undesirable adaptability exist against AdaSG. Fig.\ref{simi-visual}(b) shows that the samples generated for 3-bit and 5-bit precision vary greatly, while the samples from varied categories also differ greatly from each other, verifying that the samples possess desirable adaptability to varied Q, upon the fact that the category (label) information is fully exploited.

\section{Conclusion}
In this paper, we rethink date-free quantization process as a zero-sum game between two players --- a generator and a quantized network, then further develop an Adaptability-aware Sample Generation (AdaSG) method, which features a dynamic maximization-vs-minimization game process anchored on sample adaptability. The maximization process generates the sample with desirable adaptability, which is further reduced by the minimization process after recovering the performance of Q. The Balance Gap is defined to achieve the stationarity for the zero-sum game. The theoretical analysis and empirical studies validate the advantages of AdaSG to the existing arts.

\section{Acknowledgments}
This work are supported by National Natural Science Foundation of China under the grant no U21A20470, 62172136, 72188101, U1936217. Key Research and Technology Development Projects of Anhui Province (no.202004a5020043).

\bibliography{aaai23}

\end{document}